\documentclass[letterpaper]{article} 
\usepackage[draft]{aaai2027}  
\usepackage[hyphens]{url}  
\usepackage{graphicx} 
\usepackage{natbib}  
\usepackage{caption} 
\usepackage{algorithm}
\usepackage{algorithmic}

\usepackage{newfloat}
\usepackage{listings}
\DeclareCaptionStyle{ruled}{labelfont=normalfont,labelsep=colon,strut=off} 
\floatstyle{ruled}
\newfloat{listing}{tb}{lst}{}
\floatname{listing}{Listing}

\usepackage{booktabs}
\usepackage{multirow}
\usepackage{amsmath}
\usepackage{amssymb}
\usepackage{amsfonts}
\usepackage{xcolor}
\usepackage{array}
\usepackage{tabularx}
\usepackage{makecell}
\usepackage{placeins}

\newcolumntype{L}[1]{>{\raggedright\arraybackslash}p{#1}}
\newcolumntype{C}[1]{>{\centering\arraybackslash}p{#1}}
\newcolumntype{Y}{>{\centering\arraybackslash}X}

\title{Adaptive Supervised Anchoring for On-Policy Self-Distillation}
\author{
Meilin Yang\textsuperscript{*},
Zixuan Ding\textsuperscript{*,1},
Jianhao Nie,
Weite Zhang,\\
Yuxin Zhang,
Zhiming Shao,
Li Yu\textsuperscript{\ensuremath{\dagger},1},
Zhe Fu\textsuperscript{\ensuremath{\dagger},1}
}

\affiliations{}

\begin{document}

\maketitle
\begingroup
\renewcommand{\thefootnote}{}
\footnotetext{%
\textsuperscript{*}Equal contribution.
\textsuperscript{\ensuremath{\dagger}}Corresponding authors.\\
\textsuperscript{1}Renmin University of China, Beijing, China.
Correspondence to Li Yu
(\texttt{buaayuli@ruc.edu.cn})
and Zhe Fu
(\texttt{zhefu@ruc.edu.cn}).%
}
\endgroup

\begin{abstract}
On-policy self-distillation (OPSD) adapts a language model by distilling guidance from a frozen teacher on trajectories sampled from the student. Its effectiveness, however, depends critically on the quality of those trajectories. We show that when student rollouts drift from target trajectories, conditioning the teacher on off-target prefixes substantially weakens its task-relevant supervision. Controlled prefix-corruption experiments expose this failure mode, which we term rollout-conditioned signal degradation. To address this problem, we propose a unified training framework that separates two complementary supervision pathways. The first retains rollout-conditioned distribution matching, providing guidance on states the student actually visits. The second applies supervised cross-entropy on canonical ground-truth contexts, avoiding the incompatibility of imposing target tokens on erroneous rollout prefixes. Token-level rollout–target alignment is used to adapt the strength of the canonical-context anchor, emphasizing it during cold start and relaxing it as rollout quality improves. Experiments across multiple model scales, two task families, and general-reasoning benchmarks show that the proposed approach improves task acquisition over OPSD while preserving general capabilities, resulting in a more favorable empirical plasticity–stability trade-off. These findings identify context quality as a central bottleneck in on-policy self-distillation and demonstrate the value of separating rollout-conditioned guidance from canonical supervision.
\end{abstract}

\section{Introduction}

On-policy distillation (OPD) provides dense token-level supervision on
student-generated trajectories, but typically requires white-box access
to a separate, vocabulary-compatible teacher with substantial memory
and serving overhead \cite{agarwal2024gkd,minillm}.
On-policy self-distillation (OPSD) offers an appealing alternative: the
teacher and student share the same base model, while privileged context,
such as a verified solution, gives the teacher task-relevant information
unavailable to the student
\cite{zhao2026self,shenfeld2026selfdistillation}. This makes OPSD an appealing approach to acquiring new tasks without excessive loss of pretrained capabilities.

Yet this design relies on an implicit assumption: privileged task
knowledge remains accessible when the teacher is evaluated on the
student's own prefixes. This assumption is fragile for tasks whose
output schemas, action spaces, or reasoning protocols are weakly
represented in pretraining. Before acquiring these conventions, the
student can rapidly enter contexts that depart from valid trajectories.
Although the teacher still observes the target solution, that
information need not translate into target-aligned next-token guidance
under an off-target prefix.

To test this assumption, we conduct a controlled diagnostic study. The
same frozen privileged teacher achieves 80.4\% target-token accuracy
under canonical prefixes, but only 6.5\% when conditioned on natural
student rollouts before adaptation. We then progressively corrupt
canonical prefixes while keeping the problem, target solution, and
teacher fixed. Accuracy decreases monotonically to 3.8\% as corruption
increases. We refer
to this phenomenon as \emph{rollout-conditioned signal degradation}:
privileged guidance that is strongly target-aligned under canonical
prefixes can become substantially less informative on off-target student
contexts. The student-generated states that make OPSD on-policy can
therefore also weaken the supervision that distillation is intended to
provide.

A natural response is to introduce ground-truth supervision, but the two
straightforward combinations are structurally inadequate. Applying
cross-entropy directly under a diverged rollout prefix can make the
canonical target token semantically inconsistent with its conditioning
context, creating a target--context mismatch. Performing SFT before OPSD
avoids this mismatch, but only by hard-partitioning training into two
regimes: the student receives no rollout-conditioned guidance before the
switch and no canonical supervision afterward, irrespective of its
evolving rollout quality. The method consequently depends on a manually
chosen switching point. These limitations expose a deeper
\emph{supervision--context incompatibility}: rollout-conditioned
distillation requires student-generated prefixes to preserve on-policy
state coverage, whereas ground-truth supervision requires canonical
prefixes to preserve target validity. This suggests a
\emph{context-validity principle}: each supervision target should be
evaluated under a context in which it remains semantically meaningful.

\begin{figure*}[t]
    \centering
    \includegraphics[
        width=0.78\textwidth
    ]{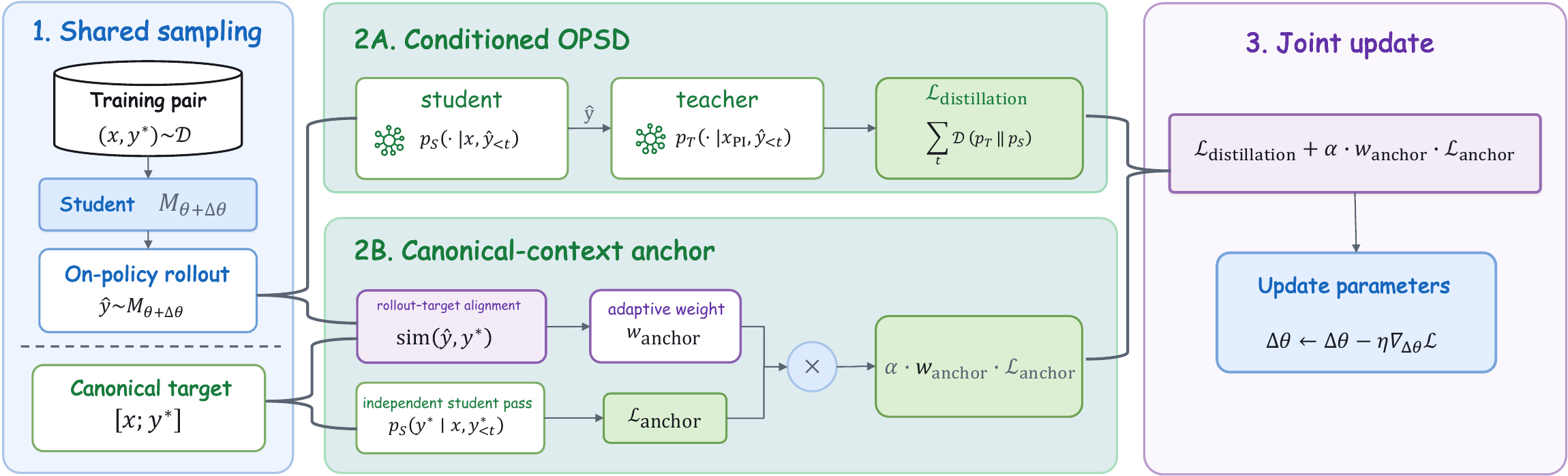}
    \caption{Overview of Supervised Distillation Steering (SDS).
    Given a training pair $(\mathbf{x},\mathbf{y}^{*})$, the student
    produces a rollout $\hat{\mathbf{y}}$. The distillation branch
    aligns the student with the privileged teacher
    $p_T(\cdot\mid\mathbf{x}_{\mathrm{PI}})$ on the rollout context,
    while the anchoring branch computes
    $\mathcal{L}_{\mathrm{anchor}}$ under the canonical prefix
    $\mathbf{y}^{*}_{<t}$. Its contribution is controlled by the
    alignment-dependent weight $w_{\mathrm{anchor}}$, and the combined
    objective updates the shared parameters $\Delta\theta$.}
    \label{fig:sds_framework}
\end{figure*}

To resolve this incompatibility, we propose
\textbf{Supervised Distillation Steering (SDS)}, a context-separated
framework that evaluates each target under a conditioning context in
which it remains semantically valid. SDS retains rollout-conditioned
self-distillation on student-generated prefixes and computes
canonical-context anchoring through an independent student forward pass.
The two branches update the same trainable parameters while keeping
their conditioning prefixes strictly separated. Crucially, SDS replaces
a fixed anchoring coefficient with an adaptive weight derived from
rollout--target token-level similarity, strengthening canonical
anchoring when alignment is poor and reducing it as alignment improves.
The anchoring contribution is thus coordinated with the student's
current behavior rather than governed by a predefined training stage.
Figure~\ref{fig:sds_framework} illustrates the framework.

We evaluate SDS across Qwen3-1.7B, Qwen3-4B, and Qwen3-8B on two
domain-adaptation tasks, ToolAlpaca and SimpleMath, and assess
general-capability retention on AIME 2024, MATH-500, and ARC-Challenge.
SDS improves domain-task acquisition over OPSD while retaining more
general capability than SFT. Training dynamics and comparisons with
rollout-context cross-entropy, fixed-weight variants, and staged
SFT-to-OPSD training further support context separation and adaptive
anchoring.

The main contributions are:
\begin{itemize}
    \item We identify \emph{rollout-conditioned signal degradation}, a
    context-dependent failure mode in privileged on-policy
    self-distillation, and link it to prefix quality through natural
    rollouts and controlled prefix corruption.

    \item We formulate context-valid supervision as a design principle
    and propose \textbf{Supervised Distillation Steering (SDS)}, which
    combines context-separated supervision with alignment-dependent
    adaptive anchoring.

    \item Experiments across three model scales demonstrate improved
    acquisition on two domain tasks and stronger general-capability
    retention than SFT. Training-dynamics analyses and ablations further
    support the two key design components.
\end{itemize}
\section{Related Work}

\subsection{On-Policy Distillation and Self-Distillation}

Knowledge distillation~\cite{hinton2015distilling,gou2021knowledge}
transfers knowledge from a teacher to a student model. In the LLM era,
distillation has further incorporated teacher-generated rationales as
additional supervision~\cite{hsieh2023distilling}. OPD instead trains the student on its own generations rather
than static teacher demonstrations. GKD~\cite{agarwal2024gkd} matches
token-level teacher distributions on student rollouts, while
MiniLLM~\cite{minillm} optimizes a reverse-KL objective on
student-sampled text. Self-distillation further shares the same base
model between teacher and student. SDFT~\cite{shenfeld2026selfdistillation}
and OPSD~\cite{zhao2026self} provide the teacher with privileged
context---expert demonstrations and verified reasoning traces,
respectively---while the student observes only the original question
and learns on-policy from the teacher's distribution.
This privileged design ties the teacher signal to the context it is queried under, and a pattern mismatch between teacher and student responses can miscalibrate token-level supervision and destabilize training \cite{yang2026ogls}. Prior remedies correct this signal within the rollout context, for example by steering the teacher logits; we instead separate supervision so that it is evaluated under the context in which it remains valid, and set its weight online from token-level sequence similarity between rollout and target rather than following a preset schedule.

\subsection{Capability Preservation in Post-Training}

Task-specific adaptation risks eroding general capabilities, a degradation classically framed as catastrophic forgetting \cite{kirkpatrick2017overcoming} and documented for large language models under continual fine-tuning \cite{luo2023empirical}. Recent work has further examined post-training forgetting at the level of individual pretrained capabilities, showing that aggregate retention scores can obscure substantial variation across capabilities and training stages \cite{harmon2026mapping}. Broader surveys categorize capability-preserving adaptation methods into rehearsal-, regularization-, and parameter-efficient approaches, underscoring the persistent plasticity--stability trade-off in continual adaptation of large language models \cite{shi2025continual}.

Among these directions, on-policy learning forgets less than supervised fine-tuning at matched performance \cite{shenfeld2025rlrazor,chen2025retaining}, low-rank adaptation limits parameter drift relative to full fine-tuning \cite{biderman2024lora}, and other approaches preserve capabilities via adapter bypassing or partial parameter freezing \cite{han2025slim,hui2025hft}. SDS instead combines rollout-conditioned self-distillation with a standard LoRA adapter, requiring no routing, freezing mechanism, or separate retention dataset.

\section{Method}
We first formalize OPSD and introduce the notation used throughout this section. Because OPSD evaluates teacher supervision on student-generated prefixes, its learning signal may become unreliable as the rollout departs from a valid solution trajectory. We therefore organize our analysis around three research questions that progressively examine whether the teacher signal remains informative, why it degrades, and whether direct cross-entropy supervision can recover the missing task signal. This diagnosis motivates SDS, which separates rollout-conditioned distillation from canonical-context supervision.

\subsection{Preliminaries}
\subsubsection{On-Policy Self-Distillation}

OPSD instantiates both teacher and student from the same base model,
but decouples them along two axes: conditioning context and parameter
update.  Given a task-adaptation dataset
$\mathcal{S}=\{(\mathbf{x}_i,\,\mathbf{y}_i^\star)\}_{i=1}^N$, let
the \emph{privileged input}
$\mathbf{x}_{\mathrm{PI}} = [\mathbf{x};\,\mathbf{y}^\star]$ denote
the concatenation of the problem and its ground-truth solution.  Let
$\theta_0$ denote the initial parameters at the start of training.  The
\emph{teacher policy}
$p_T(\cdot\mid\mathbf{x}_{\mathrm{PI}})
  \triangleq p_{\theta_0}(\cdot\mid\mathbf{x}_{\mathrm{PI}})$
is frozen at $\theta_0$ throughout training and conditions on this
privileged input, while the \emph{student policy}
$p_S(\cdot\mid\mathbf{x})
  \triangleq p_{\theta_0+\Delta\theta}(\cdot\mid\mathbf{x})$
carries the trainable increment $\Delta\theta$, which is updated
continuously during training, and observes only the problem.  Because
the teacher sees the target solution, its next-token distribution is
sharper and better aligned with correct trajectories.

At each step the student generates an on-policy rollout
$\hat{\mathbf{y}}\sim p_S(\cdot\mid\mathbf{x})$.  Both policies
evaluate this rollout: at position~$t$ they produce next-token
distributions
$p_T^t = p_T(\cdot\mid\mathbf{x}_{\mathrm{PI}},\,
  \hat{\mathbf{y}}_{<t})$
and
$p_S^t = p_S(\cdot\mid\mathbf{x},\,\hat{\mathbf{y}}_{<t})$
over the same student-generated prefix.  The OPSD objective minimises
the trajectory-averaged per-token divergence between these two
distributions, providing dense supervision at every token position on
the student's own behaviour distribution:
\begin{equation}
  \mathcal{L}_{\mathrm{OPSD}}
  = \frac{1}{|\hat{\mathbf{y}}|}
    \sum_{t=1}^{|\hat{\mathbf{y}}|}
    \mathcal{D}\!\bigl(p_T^t \,\|\, p_S^t\bigr),
  \label{eq:opsd_loss}
\end{equation}
where $\mathcal{D}$ is an $f$-divergence such as forward KL or the
generalised Jensen--Shannon divergence.  Gradients propagate only
through the student's logits $p_{\theta_0+\Delta\theta}$; the teacher
$p_{\theta_0}$ remains fixed at the initial checkpoint throughout
training and never receives gradient updates.

With the OPSD formulation in place, we now examine the reliability of its rollout-conditioned teacher signal.

\subsection{RQ1: Does the Teacher Remain Informative Under Student Rollouts?}
\label{sec:fidelity}

Equation~\ref{eq:opsd_loss} relies on the teacher distribution
$p_T(\cdot \mid \mathbf{x}_{\mathrm{PI}}, \hat{\mathbf{y}}_{<t})$
remaining informative under student-generated prefixes. We examine whether this condition holds before the student has adapted to the target task.

To quantify the available teacher signal, we define the target-token accuracy under a context sequence $c$ as
\begin{equation}
\mathrm{Acc}_T(c)
=
\mathbb{E}_{(\mathbf{x},\mathbf{y}^*),t}
\left[
\mathbf{1}
\left\{
\arg\max_v
p_T(v \mid \mathbf{x}_{\mathrm{PI}},c_{<t})
=
y_t^*
\right\}
\right],
\end{equation}
where $c=\mathbf{y}^*$ uses canonical prefixes and
$c=\hat{\mathbf{y}}$ uses student-rollout prefixes. We further define the signal-retention ratio as
\begin{equation}
\rho(\hat{\mathbf{y}})
=
\frac{\mathrm{Acc}_T(\hat{\mathbf{y}})}
{\mathrm{Acc}_T(\mathbf{y}^*)}.
\end{equation}

\paragraph{Empirical observation.}
We evaluate Qwen3-1.7B and Qwen3-4B on a fixed set of 200 SimpleMath examples before task-specific training. Under canonical prefixes, the teacher achieves target-token accuracies of 78.0\% and 80.4\%, respectively. In contrast, the corresponding student rollouts attain only 2.5\% and 1.0\% exact-match accuracy. Conditioning the teacher on these rollouts reduces its target-token accuracy to 3.0\% and 6.5\%, yielding retention ratios of 3.8\% and 8.1\%.

Together, these results show that a teacher that is accurate under canonical contexts may provide little task-relevant supervision under student-generated contexts. We refer to this loss as \emph{rollout-conditioned signal degradation}.

\subsection{RQ2: Does Context Degradation Cause Teacher Signal Loss?}

\label{sec:root_cause}

The natural-rollout results in RQ1 show that teacher accuracy collapses under student-generated contexts, but they conflate student policy quality with the teacher's sensitivity to its conditioning context. To disentangle these factors, we conduct two complementary analyses: a controlled \emph{prefix corruption} experiment that probes the effect of context degradation, and a \emph{real-rollout binning} analysis that examines whether the same relationship appears under naturally generated prefixes. The controlled experiment enables us to vary prefix quality while holding the input and target trajectory fixed, thereby isolating the effect of context degradation on the teacher signal. However, synthetic corruption may introduce error patterns that differ from those produced by an actual student policy. The rollout-binning analysis therefore provides a complementary view by testing whether teacher signal quality varies systematically with trajectory quality within real student rollouts. Together, these analyses determine whether the degradation observed in RQ1 reflects genuine sensitivity to conditioning context rather than an artifact of either poor overall student performance or the chosen corruption procedure.

\paragraph{Controlled prefix corruption.}
Starting from a ground-truth trajectory $\mathbf{y}^*$, we randomly
replace a fraction $\epsilon\in[0,1]$ of its prefix tokens with
uniformly sampled vocabulary items, yielding a corrupted context
$\tilde{\mathbf{y}}(\epsilon)$. We measure teacher target-token
accuracy as a proxy for task-relevant signal fidelity.

\begin{figure}[t]
    \centering
    \includegraphics[width=\columnwidth]
        {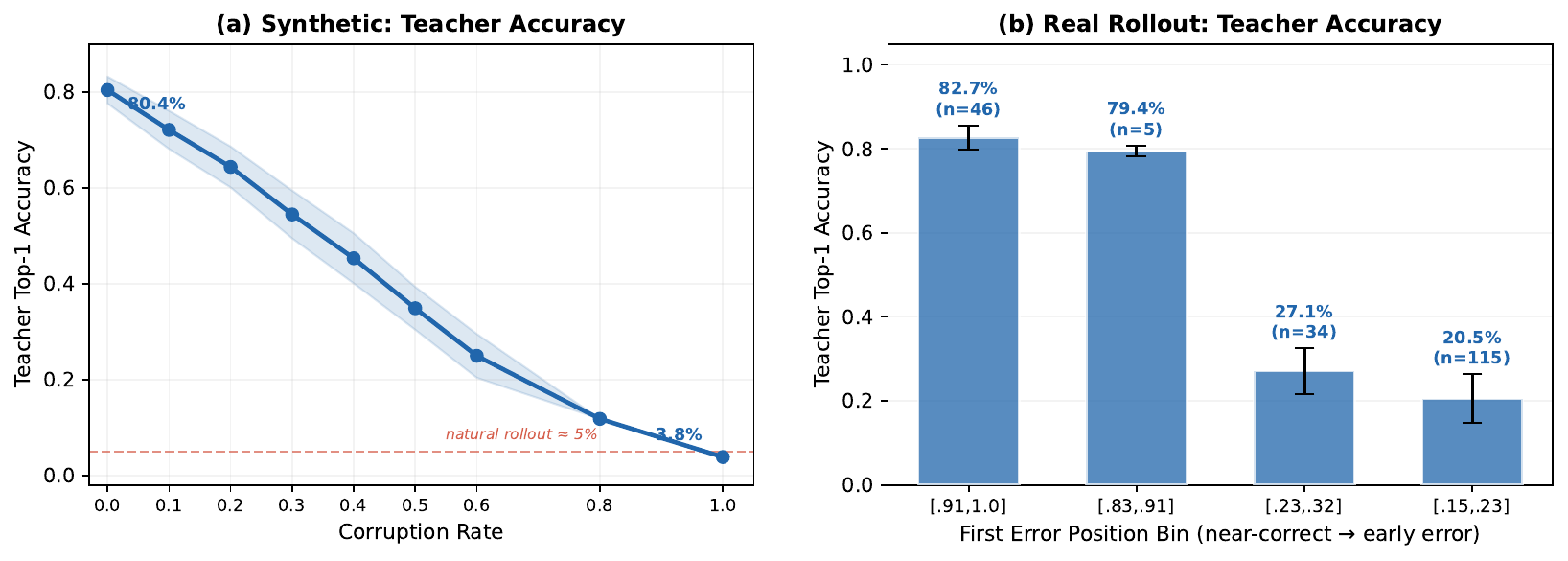}
    \caption{Teacher target-token accuracy under controlled and natural
    context degradation on Qwen3-4B SimpleMath. \textbf{(a)} Accuracy
    decreases from 80.4\% to 3.8\% as synthetic corruption increases.
    \textbf{(b)} Post-divergence accuracy increases from 20.5\% for
    early-error rollouts to 82.7\% for nearly correct rollouts. Error
    bars show within-bin standard deviation.}
    \label{fig:controlled_corruption}
\end{figure}

Figure~\ref{fig:controlled_corruption}(a) reports results on Qwen3-4B
using 200 SimpleMath samples. Teacher accuracy decreases monotonically
with the corruption rate, from 80.4\% at $\epsilon=0$ to 3.8\% at
$\epsilon=1$. The fully corrupted setting is close to the approximately
5\% accuracy observed under natural cold-start rollouts (dashed line).
These results show that the teacher's ability to predict target tokens
is highly sensitive to the quality of its conditioning prefix.

\paragraph{Validation on real student rollouts.}
To examine whether this relationship also appears under naturally
generated contexts, we collect 200 rollouts from an SDS student at
training step~50 and bin them by \emph{first error position}, defined
as the normalized location of the earliest token-level divergence
from the ground-truth trajectory. To avoid mechanically crediting
nearly correct rollouts for their longer correct prefixes, teacher
accuracy is computed only over target positions following the first
divergence.

Figure~\ref{fig:controlled_corruption}(b) shows a consistent monotonic
association. Teacher accuracy is 20.5\% for early-diverging rollouts
(bin $[0.15,0.23]$) and increases to 82.7\% for nearly correct
rollouts (bin $[0.91,1.0]$). Thus, the relationship observed under
controlled corruption is also present in natural student-generated
contexts. Together, the two analyses provide evidence that degraded
prefixes reduce the accessibility of task-relevant teacher knowledge,
rather than the controlled result being solely an artifact of random
token corruption.

\subsection{RQ3: Is Naive Cross-Entropy Injection Sufficient?}
\label{sec:naive_injection}

The analysis above shows that task-relevant teacher fidelity deteriorates
under degraded contexts. A natural remedy is therefore to supplement the
OPSD objective in Eq.~\eqref{eq:opsd_loss} with a ground-truth
cross-entropy term, thereby directly injecting task-relevant supervision:

\begin{gather}
  \mathcal{L}_{\mathrm{CE}}^{\mathrm{naive}}
  =
  -\frac{1}{|\mathbf{y}^*|}
  \sum_{t=1}^{|\mathbf{y}^*|}
  \log p_S\!\left(
    y_t^* \mid \mathbf{x}, \hat{\mathbf{y}}_{<t}
  \right),
  \label{eq:naive_ce}
  \\
  \mathcal{L}_{\mathrm{OPSD+CE}}^{\mathrm{naive}}
  =
  \mathcal{L}_{\mathrm{OPSD}}
  +
  \lambda_{\mathrm{CE}}
  \mathcal{L}_{\mathrm{CE}}^{\mathrm{naive}},
  \label{eq:naive_objective}
\end{gather}

where $\lambda_{\mathrm{CE}}$ controls the contribution of direct
ground-truth supervision. Note that the OPSD term $\mathcal{L}_{\mathrm{OPSD}}$
is averaged over the aligned token span $T$ where distillation is evaluated,
while the cross-entropy term $\mathcal{L}_{\mathrm{CE}}^{\mathrm{naive}}$ is
averaged over the full output sequence length $|\mathbf{y}^*|$.

However, this construction introduces a subtle but critical semantic
mismatch. The supervised target $y_t^*$ is evaluated under the student's
rollout prefix $\hat{\mathbf{y}}_{<t}$ rather than under its canonical
prefix $\mathbf{y}^*_{<t}$. When the rollout has already diverged,
$y_t^*$ is no longer necessarily the appropriate next token for that
prefix. The cross-entropy loss therefore optimizes for a target that is
semantically inconsistent with the conditioning context and can provide
an unreliable learning signal during early training. Consistent with
this concern, Naive OPSD+CE achieves only 55.20\% accuracy on
SimpleMath and its AIME general-reasoning performance drops to 3.33\%
in Table~\ref{tab:ablation}. These outcomes motivate context separation
but do not by themselves identify a gradient-conflict mechanism.

These findings motivate SDS, which evaluates rollout-conditioned
distillation and ground-truth anchoring under separate, semantically
valid contexts.

\subsection{Supervised Distillation Steering}

We propose \textbf{Supervised Distillation Steering} (SDS), a context-separated training objective. SDS combines a rollout-conditioned distillation term with a ground-truth-conditioned anchoring term. The two terms share the same trainable student parameters $\Delta\theta$, while their conditioning contexts are kept strictly separate. The distillation term aligns the student with the teacher under the rollout context $[\mathbf{x}; \hat{\mathbf{y}}]$, whereas the anchoring term provides target-based guidance under the ground-truth context $[\mathbf{x}; \mathbf{y}^*]$. The strength of the anchoring term is adaptively determined by the similarity between the current rollout and the target trajectory.

\subsubsection{Context-Separated Anchoring}

The anchoring branch evaluates a teacher-forcing cross-entropy loss on the ground-truth prefix. Given input $\mathbf{x}$ and canonical trajectory $\mathbf{y}^*$:

\begin{equation}
    \mathcal{L}_{\text{anchor}}
    =
    -\frac{1}{|\mathbf{y}^*|}
    \sum_{t=1}^{|\mathbf{y}^*|}
    \log p_S
    \left(
        y^*_t
        \mid
        \mathbf{x},
        \mathbf{y}^*_{<t}
    \right).
    \label{eq:anchor}
\end{equation}

The context separation is enforced at the forward-pass level: this loss is computed on $[\mathbf{x}; \mathbf{y}^*]$, while the distillation loss uses $[\mathbf{x}; \hat{\mathbf{y}}]$. The two branches share the same trainable parameters $\Delta\theta$; their conditioning contexts remain separate. This design directly addresses the mismatch identified in Eq.~\ref{eq:naive_ce}, where the target token $y^*_t$ is scored under a diverged rollout prefix $\hat{y}_{<t}$. By evaluating $y^*_t$ under its own canonical prefix $y^*_{<t}$, the anchoring gradient operates in a context where the target token is semantically valid.

We use full separation rather than partial alternatives such as selective token masking or mixed prefixes. The controlled perturbation results in Figure~\ref{fig:controlled_corruption}(a) show that teacher accuracy degrades monotonically with prefix corruption ratio, so any context that retains rollout-prefix components still carries degraded signal. Full separation avoids this by construction.

\subsubsection{Adaptive Anchoring Weight}

The anchoring branch should contribute more when student rollouts diverge from the target and less as alignment improves. We define $\mathrm{sim}(\hat{\mathbf{y}}, \mathbf{y}^*)$ as a lightweight token-level similarity that quantifies the overlap between rollout and target sequences. Any graded sequence-level similarity can fill this role; our default instantiation and an ablation over alternatives will be given in Appendix. The adaptive anchoring weight is $w_{\text{anchor}} = 1 - \mathrm{sim}(\hat{\mathbf{y}},\, \mathbf{y}^*)$. The full SDS objective at the token level is:
\begin{multline}
\label{eq:sds}
\mathcal{L}_{\mathrm{SDS}}
\;=\;
\frac{1}{|\hat{\mathbf{y}}|}
\sum_{t=1}^{|\hat{\mathbf{y}}|}
\mathcal{D}\!\left(
    p_T\!\left(\cdot \mid \mathbf{x}_{\mathrm{PI}},\,
    \hat{\mathbf{y}}_{<t}\right)
    \;\middle\|\;
    p_S\!\left(\cdot \mid \mathbf{x},\,
    \hat{\mathbf{y}}_{<t}\right)
\right)
\\
+\;
\alpha \, w_{\mathrm{anchor}}
\left(
    -\frac{1}{|\mathbf{y}^*|}
    \sum_{t=1}^{|\mathbf{y}^*|}
    \log p_S\!\left(
        y_t^* \mid \mathbf{x},\,
        \mathbf{y}_{<t}^*
    \right)
\right).
\end{multline}

where $\alpha$ controls the maximum contribution of the anchoring term. Following OPSD, we instantiate $\mathcal{D}$ as the generalized Jensen--Shannon divergence. For a mixing weight $\beta \in [0,1]$, let
$m_\beta = \beta p_T + (1-\beta)p_S$. The divergence is defined as
\begin{equation}
\begin{aligned}
\operatorname{JSD}_\beta(p_T \| p_S)
&= \beta D_{\mathrm{KL}}(p_T \| m_\beta) \\
&\quad + (1-\beta)
D_{\mathrm{KL}}(p_S \| m_\beta).
\end{aligned}
\label{eq:jsd}
\end{equation}
The generalized Jensen--Shannon divergence is bounded, preventing the loss from growing unbounded when the teacher and student distributions differ substantially. Importantly, the two terms in Eq.~\ref{eq:sds} are evaluated under distinct prefixes: $\hat{y}_{<t}$ for rollout-conditioned distillation and $y^*_{<t}$ for ground-truth supervision.

Since $w_{\text{anchor}}$ is computed from the current policy's rollout
at each step, the anchoring strength adjusts automatically as training
progresses shown in Figure~\ref{fig:training_dynamics_8b}.
\section{Experiments}

\subsection{Setup}

\paragraph{Models and datasets.}
We select Qwen3 for its open weights, which enable reproducible
self-distillation unlike proprietary systems such as GPT-4
\cite{openai2023gpt4}, and its competitive performance relative to
similarly sized open-weight models such as Llama
\cite{llamateam2024llama3,qwen3}.We evaluate its 1.7B, 4B, and 8B variants on two
adaptation tasks with distinct output structures: the SDFT-processed
version of \textbf{ToolAlpaca}
\cite{tang2023toolalpaca,shenfeld2026selfdistillation}, which requires
multi-turn structured tool calls, and \textbf{SimpleMath}
\cite{simplemath}, which requires decimal arithmetic under a prescribed
reasoning template and fixed four-decimal answer format. Task acquisition
is measured on the held-out test set of each adaptation task. We assess
transfer-side capability retention on AIME 2024 \cite{aime2024},
MATH-500 \cite{lightman2024lets}, and ARC-Challenge
\cite{clark2018think}, none of which is used for task-specific training.

\paragraph{Baselines.} We compare four primary methods, all trained on the same dataset: (1) \textbf{SFT}, standard supervised fine-tuning with cross-entropy loss; (2) \textbf{GRPO} \cite{shao2024deepseekmath}, group relative policy optimization with binary outcome rewards; (3) \textbf{OPSD} \cite{zhao2026self}, on-policy self-distillation with privileged teacher guidance; and (4) \textbf{SDS} (ours), OPSD with supervised distillation steering. We further compare SDS with naive OPSD+CE, fixed-weight hybrids, and two-stage SFT$\to$OPSD training to isolate the effects of context separation, adaptive weighting, and endogenous scheduling (see the Ablation Studies section).

\paragraph{Implementation details.} All methods use LoRA \cite{hu2022lora} applied to all attention and MLP modules. For OPSD and SDS, the teacher is served via colocated vLLM. All reported results are averaged over three random seeds to reduce variance. Full hyperparameters and evaluation configurations will be provided in Appendix.

\subsection{Main Results}
\label{sec:main_results}

\subsubsection{Task Acquisition}
Table~\ref{tab:model-size-method-results} presents results on the task acquisition benchmarks across model sizes.

\begin{table*}[t]
    \centering
    \footnotesize
    \setlength{\tabcolsep}{3.5pt}
    \renewcommand{\arraystretch}{0.78}

    \begin{tabular}{@{}llccccccc@{}}
        \toprule
        \multirow{2}{*}{Model Size}
        & \multirow{2}{*}{Method}
        & \multicolumn{2}{c}{Tools}
        & \multicolumn{5}{c}{SimpleMath} \\
        \cmidrule(lr){3-4}
        \cmidrule(lr){5-9}
        &
        & Pass@12 $\uparrow$
        & Turn Acc. $\uparrow$
        & Overall Acc. $\uparrow$
        & Add. $(+)$ $\uparrow$
        & Sub. $(-)$ $\uparrow$
        & Mul. $(\times)$ $\uparrow$
        & Div. $(\div)$ $\uparrow$ \\
        \midrule

        \multirow{5}{*}{Qwen3-1.7B}
        & Base
        & 61.76 & 40.65 & 2.93
        & 5.20 & 5.60 & 0.53 & 0.40 \\
        & SFT
        & 69.61 & 48.41 & 51.20
        & 98.40 & 99.20 & 0.80 & 6.40 \\
        & GRPO
        & 67.65 & 46.44 & 44.50
        & 90.00 & 86.13 & 1.33 & 0.40 \\
        & OPSD
        & 67.16 & 44.53 & 17.80
        & 30.80 & 36.80 & 1.20 & 2.40 \\
        & SDS
        & 68.63 & 51.36 & 51.00
        & 97.20 & 98.00 & 0.80 & 8.00 \\
        \midrule

        \multirow{5}{*}{Qwen3-4B}
        & Base
        & 61.76 & 46.15 & 3.77
        & 7.60 & 4.53 & 2.53 & 0.40 \\
        & SFT
        & 70.10 & 50.00 & 56.43
        & 100.00 & 100.00 & 1.20 & 24.40 \\
        & GRPO
        & 64.22 & 46.32 & 55.20
        & 92.40 & 95.20 & 15.60 & 17.60 \\
        & OPSD
        & 62.75 & 46.99 & 23.23
        & 44.53 & 44.40 & 3.20 & 0.80 \\
        & SDS
        & 72.06 & 51.01 & 62.43
        & 99.20 & 98.80 & 25.20 & 26.53 \\
        \midrule

        \multirow{5}{*}{Qwen3-8B}
        & Base
        & 69.12 & 49.91 & 2.13
        & 4.67 & 3.20 & 0.40 & 0.27 \\
        & SFT
        & 76.47 & 49.45 & 55.57
        & 100.00 & 100.00 & 0.80 & 21.47 \\
        & GRPO
        & 67.16 & 49.62 & 43.07
        & 82.13 & 86.80 & 1.33 & 2.00 \\
        & OPSD
        & 68.63 & 50.29 & 10.33
        & 26.80 & 11.60 & 1.60 & 1.20 \\
        & SDS
        & 70.59 & 53.10 & 56.97
        & 99.60 & 100.00 & 1.60 & 26.53 \\
        \bottomrule
    \end{tabular}
    \caption{Task acquisition across model sizes. SDS substantially
    improves over OPSD on SimpleMath and remains competitive with SFT.}
    \label{tab:model-size-method-results}
\end{table*}

SDS substantially improves task acquisition over OPSD across all three model sizes, especially on SimpleMath. On the Qwen3-4B model, SDS achieves 72.06\% Pass@12 on Tools (vs.\ 70.10\% for SFT and 62.75\% for OPSD) and 62.43\% overall accuracy on SimpleMath (vs.\ 56.43\% for SFT and 23.23\% for OPSD). On Tools, SFT remains competitive at 1.7B and 8B, where its direct supervision on expert trajectories is sufficient and capability erosion is less critical. However, SDS achieves the best or near-best SimpleMath performance across all scales, and its advantage grows with task difficulty (e.g., multiplication and division). Combined with its substantially better general capability retention (Table~\ref{tab:aime24-math500-4b}), SDS offers a more favorable trade-off than SFT for scenarios where both task performance and general capability matter.

\subsubsection{General Capability Preservation}

Table~\ref{tab:aime24-math500-4b} presents results on mathematical and science-reasoning benchmarks for the Qwen3-4B model.

\begin{table}[!htbp]
  \centering
  \scriptsize
  \setlength{\tabcolsep}{2.2pt}
  \begin{tabular}{lcccccccc}
  \toprule
  \multirow{2}{*}{Method}
  & \multicolumn{3}{c}{AIME24}
  & \multicolumn{3}{c}{MATH-500}
  & \multicolumn{2}{c}{ARC-Challenge} \\
  \cmidrule(lr){2-4} \cmidrule(lr){5-7} \cmidrule(lr){8-9}
  & Avg@8 & Pass@8 & Maj@8
  & Avg@8 & Pass@8 & Maj@8
  & Acc. & Format \\
  \midrule
  Base
  & 72.92 & 80.00 & 80.00
  & 95.63 & 98.40 & 97.00
  & 92.15 & 98.12 \\
  SFT
  & 38.33 & 80.00 & 73.33
  & 56.00 & 84.60 & 80.60
  & 66.13 & 73.29 \\
  GRPO
  & 61.25 & 76.67 & 70.00
  & 93.75 & 97.60 & 96.00
  & 90.58 & 96.63 \\
  OPSD
  & 71.25 & 86.67 & 80.00
  & 94.90 & 98.20 & 96.00
  & 91.30 & 97.10 \\
  SDS
  & 67.98 & 86.67 & 83.33
  & 93.65 & 98.20 & 95.00
  & 87.75 & 95.69 \\
  \bottomrule
  \end{tabular}
  \caption{Capability preservation on Qwen3-4B after SimpleMath adaptation.}
  \label{tab:aime24-math500-4b}
  \end{table}

SFT substantially degrades all three transfer-side probes: AIME24 Avg@8
drops from 72.92\% to 38.33\%, MATH-500 Avg@8 from 95.63\% to
56.00\%, and ARC-Challenge accuracy from 92.15\% to 66.13\%; format
compliance also falls from 98.12\% to 73.29\%.

OPSD preserves the evaluated capabilities well but remains weak at task
acquisition. SDS occupies the middle ground, retaining 93.2\% of base
AIME24 Avg@8 and 95.2\% of ARC-Challenge accuracy while achieving the
best Qwen3-4B task performance in Table~\ref{tab:model-size-method-results}.
In comparison, SFT retains only 52.6\% of base AIME24 Avg@8 and 71.8\%
of base ARC-Challenge accuracy.

MATH-500 differences are smaller, with all methods except SFT retaining
over 93\% Avg@8. On ARC-Challenge, SDS remains below OPSD and GRPO but
substantially outperforms SFT and preserves a 95.69\% format rate.
Appendix will report the corresponding
cross-scale retention results. Pure OPSD remains the strongest
preservation-oriented baseline, whereas SDS provides substantially
stronger task acquisition.

\subsubsection{The Pareto Trade-off}

\begin{figure}[!b]
    \centering
    \includegraphics[
        width=0.6\columnwidth
    ]{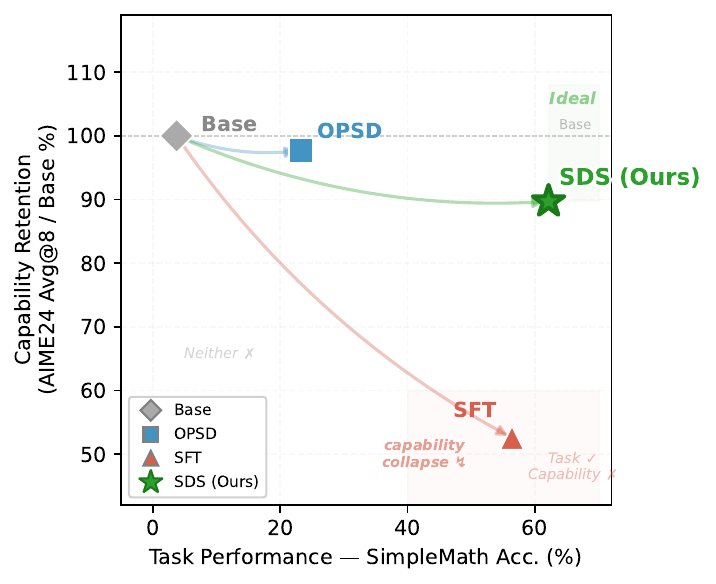}
    \caption{Task acquisition versus AIME24 capability retention on
    Qwen3-4B. SDS achieves 62.43\% SimpleMath accuracy while retaining
    93.2\% of the base AIME24 Avg@8, providing a favorable empirical
    plasticity--stability trade-off.}
    \label{fig:pareto}
\end{figure}

Figure~\ref{fig:pareto} summarizes the empirical
plasticity--stability trade-off. Starting from the same base model, SFT
acquires the target task but substantially degrades AIME24 performance,
whereas OPSD preserves AIME24 but learns SimpleMath less effectively.
SDS achieves a stronger balance by separating the two objectives:
rollout-conditioned distillation maintains distributional continuity,
while canonical-context anchoring supplies task supervision.
\subsubsection{Training Dynamics of Adaptive Anchoring}
\label{sec:training_dynamics}

\begin{figure}[t]
\centering
\includegraphics[width=0.93\columnwidth]
{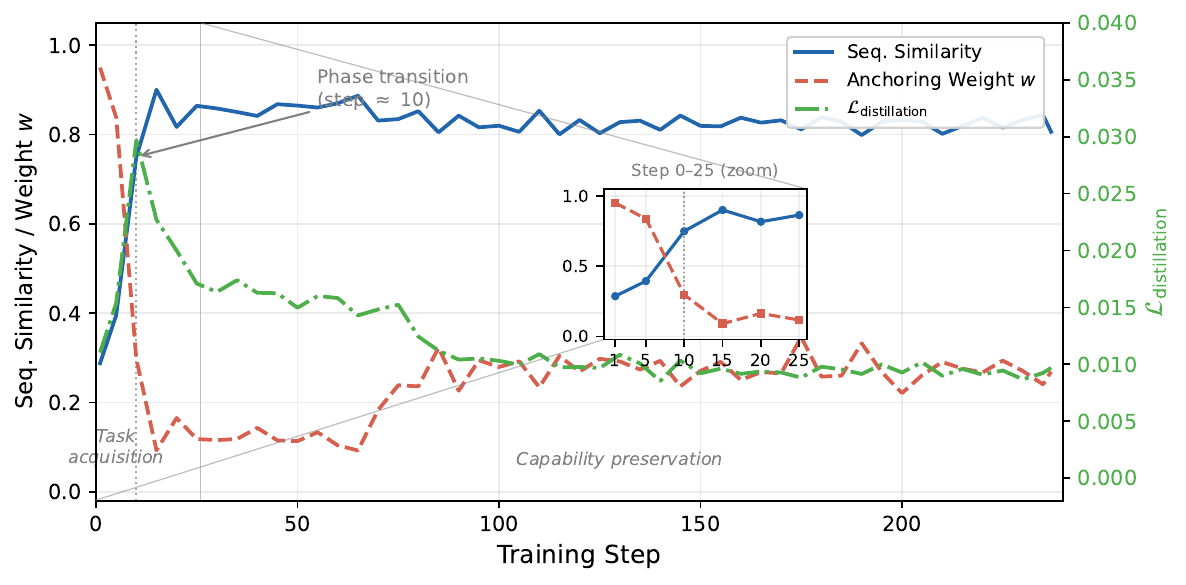}
\caption{Training dynamics of SDS on Qwen3-8B. Adaptive anchoring is
stronger early in training, decreases to a nonzero level, and
continuously coordinates with rollout-conditioned distillation without
a hard stage switch.}
\label{fig:training_dynamics_8b}
\end{figure}

To understand how SDS coordinates its two supervision branches, we
examine the dynamics of adaptive anchoring and rollout-conditioned
distillation. Figure~\ref{fig:training_dynamics_8b} shows that early
student rollouts are poorly aligned with the ground-truth trajectories,
resulting in a large $w_{\text{anchor}}$ and strong canonical anchoring.
As alignment improves, the anchoring weight decreases and the
distillation branch gains influence. The rollout-conditioned distillation
divergence initially rises during rapid adaptation and then declines as
student rollouts become more compatible with the privileged teacher.

Notably, $w_{\text{anchor}}$ does not decay to zero. It stabilizes at a
lower but nonzero level and continues to vary with rollout quality. This
behavior is consistent with a negative-feedback mechanism: anchoring is
strengthened when alignment is poor, relaxed as valid trajectories
emerge, and retained to correct subsequent deviations.

SDS therefore does not implement a hard SFT-to-OPSD switch. Instead, it
continuously balances canonical anchoring and rollout-conditioned
distillation according to the student's evolving behavior, avoiding
fixed weighting and a manually specified transition point.

\subsection{Ablation Studies}
\label{sec:ablation}

Table~\ref{tab:ablation} isolates the three design choices that distinguish SDS from simpler alternatives: context separation, adaptive weighting, and endogenous scheduling.

\begin{table}[!htbp]
    \centering
    \footnotesize
    \setlength{\tabcolsep}{3.5pt}
    \renewcommand{\arraystretch}{0.95}

    \begin{tabular}{@{}llcc@{}}
        \toprule
        Variant & Control & Acc. & AIME-S \\
        \midrule

        SDS ($\alpha\!=\!0.2$)
        & Full
        & 62.43
        & 67.98 \\
        \midrule

        Naive OPSD+CE
        & Ctx.\ mismatch
        & 55.20
        & 3.33 \\

        Isolated SFT, $w\!=\!1$
        & Fixed weight
        & 57.20
        & 56.25 \\

        Isolated SFT, $w\!=\!0.5$
        & Fixed weight
        & 35.40
        & 72.48 \\

        SFT$\to$OPSD
        & Manual sched.
        & 55.03
        & 35.42 \\

        \bottomrule
    \end{tabular}
    \caption{SDS ablations on Qwen3-4B SimpleMath. AIME-S is AIME24 Avg@8.}
    \label{tab:ablation}
\end{table}
\paragraph{Context isolation.}
Naive OPSD+CE injects ground-truth cross-entropy directly into the rollout-conditioned computation graph, precisely the target--context mismatch analyzed in the RQ3 analysis. It achieves only 55.20\% accuracy and 3.33\% AIME Avg@8 --- the latter representing near-total collapse of general reasoning. By contrast, SDS computes the anchoring loss under canonical prefixes and reaches 62.43\% accuracy with 67.98\% AIME Avg@8. This confirms that context validity, not supervision quantity, is the fundamental bottleneck: the same ground-truth signal that helps under canonical prefixes actively harms under mismatched rollout prefixes.

\paragraph{Adaptive weighting.}
Fixed-weight isolated SFT removes the closed-loop schedule while preserving context isolation. With $w\!=\!1$ (always-on anchoring), task accuracy reaches 57.20\% but AIME degrades to 56.25\%, indicating that persistent hard supervision erodes general capability. With $w\!=\!0.5$, task accuracy collapses to 35.40\% while AIME rises to 72.48\%, showing that insufficient anchoring fails to guide task acquisition. No fixed weight simultaneously achieves strong task performance and capability preservation, confirming that the alignment-aware schedule is essential for navigating the trade-off.

\paragraph{Endogenous vs.\ manual scheduling.}
SFT$\to$OPSD uses a two-stage pipeline with a manual transition, providing the same supervision signals as SDS but in fragmented stages. It achieves 55.03\% accuracy and 35.42\% AIME Avg@8, underperforming SDS on both axes. This demonstrates that fragmenting the supervision signal across stages introduces a discontinuity that the endogenous, closed-loop transition in SDS avoids.


\section{Analysis and Discussion}

\paragraph{Why context isolation works.}
The ablation results reveal a principle that, to our knowledge, has not been explicitly articulated in the distillation literature: \emph{where} supervision is computed matters more than \emph{how much} is applied. Naive OPSD+CE and SDS use identical ground-truth signals and identical loss functions, yet differ drastically in outcome because one evaluates targets under mismatched rollout prefixes and the other under canonical prefixes. This suggests that future work on combining on-policy and off-policy signals should consider context compatibility as a design constraint alongside weighting and scheduling.

\paragraph{Choice of alignment metric.}
The closed-loop anchoring weight depends on the distance between student
rollouts and ground-truth targets. Appendix will
compare four alternatives. SDS remains stable across the tested graded
metrics, with limited differences in AIME 2024 retention, whereas binary
exact match is less effective on multiplication.

\paragraph{Role of $\alpha$ in the plasticity--stability trade-off.}
The scaling hyperparameter $\alpha$ controls the maximum influence of the anchoring branch and governs where SDS operates on the plasticity--stability plane. On Qwen3-4B SimpleMath, $\alpha=0.2$ yields the best overall balance (62.43\% task accuracy, 93.2\% AIME retention); $\alpha=0.1$ shifts toward capability preservation (56.73\% task accuracy, 97.7\% retention), while $\alpha=0.5$ shifts toward task acquisition without surpassing $\alpha=0.2$ on accuracy (57.13\%, 82.8\% retention). This mirrors the fixed-weight ablation, where always-on anchoring ($w\!=\!1$) reaches only 57.20\%: excessive anchoring suppresses the on-policy exploration that OPSD relies on, since canonical supervision dominates the update before the student explores enough of its own rollout space for teacher guidance to take effect. The adaptive weight $w_{\text{anchor}}$ relaxes anchoring as alignment improves, but a high $\alpha$ still raises supervision intensity beyond what this exploration--exploitation balance can absorb. $\alpha$ therefore acts as an interpretable knob, with lower values favoring capability retention and higher values favoring task acquisition.

\paragraph{Computational overhead.}
SDS introduces one additional student forward pass per training step due to the independent anchoring branch. Compared to two-stage pipelines (SFT warm-up followed by OPSD), SDS avoids a separate SFT phase and its associated training budget, while eliminating manual transition tuning. In practice, the overhead is approximately 30--40\% of a single OPSD step, which we consider a reasonable cost given the performance gains and the elimination of stage scheduling.


\paragraph{Limitations.}
Our evaluation is confined to a single model family, Qwen3, and two
adaptation tasks, leaving the generalization of SDS across model
families and broader task settings unverified. Moreover, controlled
prefix corruption provides only a tractable proxy for context
degradation and cannot fully reproduce the structured, history-dependent
errors that emerge in natural student rollouts. Future work should
evaluate SDS across diverse model families and develop more realistic,
semantically grounded corruption processes.


\section{Conclusion}
We studied how conditioning context affects the validity of heterogeneous
supervision during on-policy task adaptation. In OPSD with privileged
teacher context, teacher guidance degrades as student rollouts depart
from valid trajectories, revealing a mismatch between on-policy state
coverage and canonical target validity. SDS addresses this mismatch by
separating rollout-conditioned distillation from canonical anchoring and
coordinating them through rollout--target alignment. Across model scales
and tasks, SDS consistently improves task acquisition over OPSD while
preserving substantially more reasoning performance than SFT, achieving
the most favorable trade-off among the evaluated methods.

\FloatBarrier

\bibliography{aaai2027}

\FloatBarrier
\end{document}